\documentclass{article}

\usepackage[margin=1in]{geometry} 
\usepackage{lipsum}
\usepackage{enumitem}

\usepackage[utf8]{inputenc} 
\usepackage[T1]{fontenc}    
\usepackage{url}            
\usepackage{booktabs}       
\usepackage{amsfonts}       
\usepackage{nicefrac}       
\usepackage{microtype}      
\usepackage{graphicx}
\usepackage{natbib}
\usepackage{doi}
\usepackage{float}
\usepackage{amsmath}
\usepackage{amsthm}
\usepackage{multirow}
\usepackage{algorithm}
\usepackage{algpseudocode}
\usepackage{xcolor}
\usepackage{hyperref}
\hypersetup{
    colorlinks,
    linkcolor={red!50!black},
    citecolor={blue!50!black},
    urlcolor={blue!80!black}
}

\setcitestyle{authoryear, open={(},close={)}}

\newcommand{\beginsupplement}{%
        \setcounter{section}{0}
        \renewcommand{\thesection}{S\arabic{section}}%
        \renewcommand{\theHsection}{S\arabic{section}}
        \setcounter{table}{0}
        \renewcommand{\thetable}{S\arabic{table}}%
        \renewcommand{\theHtable}{S\arabic{table}}
        \setcounter{figure}{0}
        \renewcommand{\thefigure}{S\arabic{figure}}%
        \renewcommand{\theHfigure}{S\arabic{figure}}
     }

\title{\textbf{Structured Recurrent Mixers for Massively Parallelized Sequence Generation}}

    \author{\small Benjamin L. Badger \\
    \small IBM\\
    \texttt{\small ben.badger@ibm.com}
    \date{}}
    
\begin{document}
    \maketitle
    \footnotetext{The author would like to that IBM for support during research and the writing of this manuscript. Code for this paper can be found at \url{https://github.com/blbadger/mrm}}

\maketitle

\begin{abstract}\normalfont{
    Over the last two decades, language modeling has experienced a shift from the use of predominantly recurrent architectures that process tokens sequentially during training and inference to non-recurrent models that process sequence elements in parallel during training, which results in greater training efficiency and stability at the expense of lower inference throughput. Here we introduce the Structured Recurrent Mixer, an architecture that allows for algebraic conversion between a sequence parallel representation at train time and a recurrent representation at inference, notably without the need for specialized kernels or device-specific memory management. We show experimentally that this dual representation allows for greater training efficiency, higher input information capacity, and larger inference throughput and concurrency when compared to other linear complexity models. We postulate that recurrent models are poorly suited to extended sequence length scaling for information-rich inputs typical of language, but are well suited to scaling in the sample (batch) dimension due to their constant memory per sample. We provide Mojo/MAX inference implementations of SRMs exhibiting 12x the throughput and 170x the concurrency of similarly powerful Transformers inferenced on vLLM, increases characteristic of Pytorch implementations resulting in a 30\% increase in compute-constant GSM8k Pass@k. We conclude by demonstrating that SRMs are effective reinforcement learning training candidates due to this large throughput.
    }
\end{abstract}

\section{Introduction}

    Is model speed or accuracy more important for language tasks? The answer usually depends on the task, specifically whether or not it admits a way of quickly checking whether a sample is good or bad. For tasks where samples cannot be quickly verified, model accuracy is typically more important than speed, but for many tasks verification is feasible: for example, unit or functional tests may be applied to code samples, outcome evaluation to agentic workflows, self-verification to mathematical theorem proof generation via Lean. If rapid verification is possible, it is often more efficient (in terms of minimal compute spent per correct output) to generate many samples from smaller and less-capable models than fewer samples from larger, more accurate ones \citep{brown2024largelanguagemonkeysscaling, chen2025parallelscalinglawlanguage}.
    
    Most functional language benchmarks do not focus on efficiency in these terms but instead on the likelihood of a correct answer or correct next token given one or more attempts, although some more recent benchmarks have begun to incorporate efficiency in terms of monetary cost per task (for example Arc-AGI \citep{foundation2026arcagi3newchallengefrontier}). The shift away from recurrent-style architectures (recurrent neural networks \citep{rumelhart1985learning} and LSTMs \citep{hochreiter1997lstm}) occurred in the context of next token prediction accuracy as well as their relation to applied benchmarks in which model goodness metrics are based on a model's accuracy per sample regardless of the compute necessary to generate that sample. For outputs that are easily verified, a metric of accuracy per compute applied seems more appropriate, and the findings that smaller Transformer models \citep{vaswani2023attentionneed} can outperform larger ones in this context leads to the question: are architectures that perform more efficient inference than Transformers better suited to this paradigm of accuracy per compute? 

    For large-scale language modeling one usually wants to train causal models to predict each next token in a context window in parallel via token shifting, with a change in model behavior during inference to remove this sequence parallelism for efficiency so that the model predicts one token at a time with a cache storing information from previous tokens. Sequence parallel models, particularly those with $\mathcal O(n^2)$ time and $\mathcal O(n)$ space complexity for $n$ tokens, are efficient and stable to train but inefficient for sequential token generation, whereas recurrent models (in this work defined as those with $\mathcal O(n)$ time and $\mathcal O(1)$ space complexity) are efficient for sequence generation but typically do not admit easy ways to train efficiently and stably, as they are not intrinsically parallelizable over the sequence dimension. 

    As a result of these complementary deficiencies, and because of the large latency of global device memory access relative to arithmetic operations present in GPUs, there have been numerous attempts to hybridize linear-complexity architectures (Mamba SSMs for Granite 4 \citep{mishra2024granitecodemodelsfamily, ibmGraniteGranite} and Nemotron 3 \citep{nvidia2025nvidianemotron3efficient}, Gated Delta Nets for Qwen 3.5 \citep{qwen3.5} and Kimi Linear \citep{kimiteam2025kimilinearexpressiveefficient}) to transformer backbones. Hybridization, however, introduces a significant challenge: although these models gain the advantages of (smaller) variable-sized caches, they also exhibit the same drawback as Transformers with respect to cache scaling, just with lower constant values.

    These observations motivate the introduction of a new architecture, the Structured Recurrent Mixer, that we find exhibits some of the desirable information retention and training efficiency characteristics of quadratic complexity models while improving on the the efficient inference computation of current recurrent models. From an informational perspective we show that recurrent models are ill-suited to unbounded sequence length scaling but are better applied to batch dimension scaling, and show that the SRM exhibits large throughput and concurrency increases relative to other models tested without any effort into optimizing parallelization via device-specific kernels. We address the question of whether designing an architecture that exhibits efficient sample throughput properties could lead to accuracy beyond that seen for a representative Transformer, with both training and inference compute held constant. Finally we investigate the use of reinforcement learning to convert the many-sample efficiency back to single- or few-sample efficiency.

\section{Our Contribution}

    This paper details experiments testing two primary insights. The first is that masked mixers should be more easily linearizable (in the sequence dimension) than transformers due to their data-independent token mixing operations, allowing for an effective bridge between quadratic complexity and recurrent models. We find that

    \vspace{0.15cm}
    \noindent \textit{Designing recurrent model token mixing operations to capture qualitative features of trained higher-complexity models recovers some of those models' training efficiency and information retention}
    \vspace{0.15cm}

    \noindent and secondly we introduce a re-evaluation of the benefits of constant space models when applied to information-dense inputs such as language, proposing that 

    \vspace{0.15cm}
    \noindent \textit{Recurrent language models are better suited to batch (sample number) dimension rather than sequence length scaling due to their constant-sized, contiguous memory per sample}
    \vspace{0.15cm}

    \noindent for which scaling the SRMs are shown to have unparalleled throughput characteristics even among recurrent models by construction, which is desirable characteristic for any task where samples may be quickly tested. We introduce the following:

    \vspace{0.15cm}
    \begin{enumerate}[nosep]
    \item A qualitative feature mapping method from quadratic-complexity to linear-complexity models
    \item A dual-architecture sequence parallel (at train time) and batch parallel (at inference) recurrent model
    \item A Mojo/MAX inference engine for SRMs with >7x the throughput of Pytorch implementations
    \item A resampling method for applying GRPO to large batches without decreasing exploration
    \end{enumerate}

\section{Related Work}

    Recurrent neural networks were among the first models applied to the problem of variable-length sequence generation \citep{rumelhart1985learning}. The Long Short-Term Memory model, a recurrent-like model introducing multiple hidden states and associated memory streams for these sates, was introduced in order to stabilize training and better model long sequences \citep{hochreiter1997lstm}. The greater training efficiency of Transformers compared to LSTMs both with respect to loss per parameter and loss per input context length account in part for why these architectures have been largely supplanted by Transformers for most language modeling tasks \citep{kaplan2020scalinglawsneurallanguage}.

    Attempts to convert Transformers directly into recurrent-like architectures via linearized attention have experienced somewhat limited success: direct swap-ins of linear attention for dot-product attention \citep{katharopoulos2020, shen2019efficient} results in poor training efficiency and stability for causal modeling \citep{poli2023hyenahierarchylargerconvolutional}, although progress has been made by reformulating the linear attention operation. Along these lines, the RWKV architecture \citep{peng2023rwkvreinventingrnnstransformer} was introduced, and these models retain key and value vectors but perform memory mapping operations requiring the use of custom GPU kernels for efficient computation. 

    The most well-explored model architectures that allow for conversion between sequence parallel and recurrent representations exist in the state space framework, in which a state vector acts as the the recurrent hidden layer and is updated via linear operations. Refined from S2/S3 \citep{fu2023hungryhungryhipposlanguage} models, Mamba \citep{gu2024mambalineartimesequencemodeling} and Mamba 2 \citep{dao2024transformersssmsgeneralizedmodels} SSM architectures incorporate selectivity at the token level and sophisticated memory management which allows for efficient parallel scan-based state updates and thus conversion between sequence parallel and recurrent-like forms. 
    
\section{Method}

    Structured Recurrent Mixers are structured so that one is able to algebraically convert the sequence parallel representation (which uses matrix multiplication to mix token information) to the recurrent representation (which must read and write to constant-sized memory and use a fixed number of operations per token generated). Training proceeds by structuring token mixing matrices by expanding weight vectors along rows or columns, applying an optional trainable decay rate, and proceeding with shifted next token prediction.

    \subsection{Masked Mixer Parameterization}
    
    SRMs are based on the Masked Mixer, a transformer-like architecture in which attention is swapped for masked MLPs (matrix multiplications) \citep{badger2025maskedmixerslanguagegeneration}. These models apply masked matrix multiplication (or equivalently masked 1-D convolution with a stride size of 1) to the sequence dimension, which can be expressed as $Y = XM + B$ where $M \in \Bbb R^{n \times n}$ and $X, Y, B \in \Bbb R^{d \times n}$. Notably the values of $M$ are triangular masked for both training and inference, and are not data-dependent but are parameterized by trainable values in the model. These architectures exhibit similar time and space complexities to Transformers but form fewer activations ($d$ per layer) in the token mixing operation, resulting in higher throughput at a fixed number of parameters which is balanced by their lower training efficiency on a per-parameter basis.

    \subsection{Trained Masked Mixer Weights: Qualitative Feature Representation}

\begin{figure}[h]
        \centering
        \includegraphics[width=0.95\textwidth]{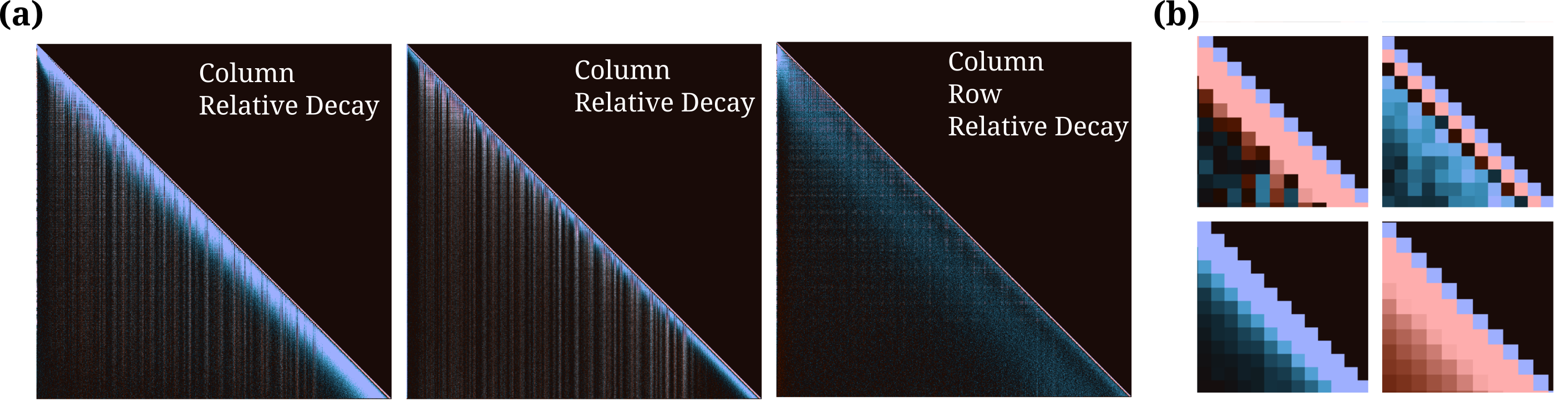}
        \caption{Causal Masked Mixer token mixing matrix weights, (a) annotated features on whole-matrix maps, (b) weights near the main diagonal. Blue indicates negative, and red positive values.}
        \label{fig1}
    \end{figure}

    Initial investigations detailed in Table \ref{tables0} demonstrated that a naive restriction of token mixing matrices to a recurrent form (row repeat matrices) results in relatively inefficient causal language model training. This motivated an analysis on the structure of trained masked mixer token mixing layers, which revealed three notable qualitative features: matrices with columns containing near-identical values, matrices with rows containing near-identical values, and matrices with weight decay with increased distance from the main diagonal (or some combination of these three features) as shown in Figure \ref{fig1}. As we explore in the next section, structuring a recurrent SRM to represent these qualitative features results in substantially higher training efficiency. Curiously, not all features of trained mixer weight matrices are apparently necessary for efficient recurrent modeling: most layers exhibit lower near-constant main diagonal weights compared to off-diagonal weights (Figures \ref{fig1}, \ref{figs1}), but introducing an independent weight for the main diagonal to SRMs does not result in significantly increased training efficiency for larger models (Table $\ref{tables12}$).

    \subsection{Architecture}

    As the token mixing layers of the SRM are restricted in order to retain a recurrent representation, we now illustrate what kind of restrictions are required. In brief, the restrictions we explore are relatively simple: rows must contain identical elements or columns must contain identical elements, and optionally a constant factor may be multiplied to each diagonal. SRMs therefore are characterized by token mixing matrices that are fully separable, with rank 1, which is why the operation may be converted into a fixed-memory recurrent operation. 
    
    To illustrate an example row repeat mixing operation, consider input $X \in \Bbb R^{d \times n}$ with $d$ hidden dimension and $n$ tokens, output $Y \in \Bbb R^{d \times n}$, mixer weight matrix $W \in \Bbb R^{n \times n}$, bias $B \in \Bbb R^{d \times n}$, and trainable decay constant $\lambda \in (0.9, 1]$. We obtain $Y$ from inputs $X, B, W$ as shown in Equation \ref{eq1} for the case where $n=3$ for simplicity. Equation \ref{eq1} may be expressed in its recurrent representation in Equation \ref{eq2}, where the recurrent hidden state (cache) is simply the sum term. 

    \begin{equation}
    \begin{pmatrix}
    \vert & \vert & \vert \\
    Y_0 & Y_1 & Y_2 \\
    \vert & \vert & \vert
    \end{pmatrix} 
    = \begin{pmatrix}
        \vert & \vert & \vert \\
        X_0 & X_1 & X_2 \\
        \vert & \vert & \vert
    \end{pmatrix} 
    \begin{pmatrix}
      \alpha_0 & \lambda \alpha_0 & \lambda^2 \alpha_0 \\
      0 & \alpha_1 & \lambda \alpha_1 \\
      0 & 0 & \alpha_2
    \end{pmatrix}
    + 
    \begin{pmatrix}
        \vert & \vert & \vert \\
        \beta_0 & \beta_1 & \beta_2 \\
        \vert & \vert & \vert
    \end{pmatrix} 
    \label{eq1}
    \end{equation}

    \begin{equation}
    \begin{split}
    Y_n &= \alpha_n X_n + \beta_n + \sum_{m=0}^{n-1} \lambda^{n-m} \alpha_m X_m 
    \end{split}
    \label{eq2}
    \end{equation}

    Similarly, a column repeat matrix multiplication with per-token biases may be reduced to the recurrent representation given in Equation \ref{eq3}, where again the sum term is the constant-size cache and $\alpha_n$ is factored out upon each update. We implement two methods for combining column and row repeat outputs: linear combinations where each head has one column and one row repeat matrix and the output is linearly combined, as well as what we term `mixed' heads where heads are evenly divided into column and row repeat layers and combined as other headed models are (via concatenation before out projection), which we term `head-parallel' layers. In either case next token prediction is efficient and requires at most two BLAS level 1 operation per recurrent layer at inference, although linearly combined layers with both row and column repeat layers require double the cache size as mixed or row- or column-only layers.
    
    We implement multiple heads per token mixing layer by training unique projections $I_h \in \Bbb R^{d_h\times d}$ for each head $0, 1, ..., h$, for head dimension $d_h = d / h$, performing token mixing on these heads independently, and projecting the concatenated outputs via $P_{out} \in \Bbb R^{d\times d}$ as shown in Equation \ref{eq10}. Equivalently, one input projection $I\in \Bbb R^{d\times d}$ may or may not be applied, heads are sliced and token mixing performed and outputs concatenated and optionally projected back.
    
    \begin{equation}
    \begin{split}
    Y = X \begin{pmatrix}
      \alpha_0 & \lambda \alpha_1 & \lambda^2\alpha_2 &  \cdots & \lambda ^{n-1}\ \alpha_n \\
      0 & \alpha_1 & \lambda \alpha_2 & \cdots & \lambda ^{n-2}\ \alpha_n \\
      0 & 0 & \alpha_2 & \cdots & \lambda^{n-3}\alpha_n \\
      \vdots & \vdots & \vdots &  \ddots & \vdots\\
      0 & 0 & 0 & \cdots & \lambda^0\alpha_n
    \end{pmatrix} + \beta  \implies  Y_n &= \alpha_n X_n + \beta_n + \alpha_n \sum_{m=0}^{m=n-1} \lambda ^{n-m}X_m 
    \end{split}
    \label{eq3}
    \end{equation}

    \begin{figure}[h]
        \centering
        \includegraphics[width=0.99\textwidth]{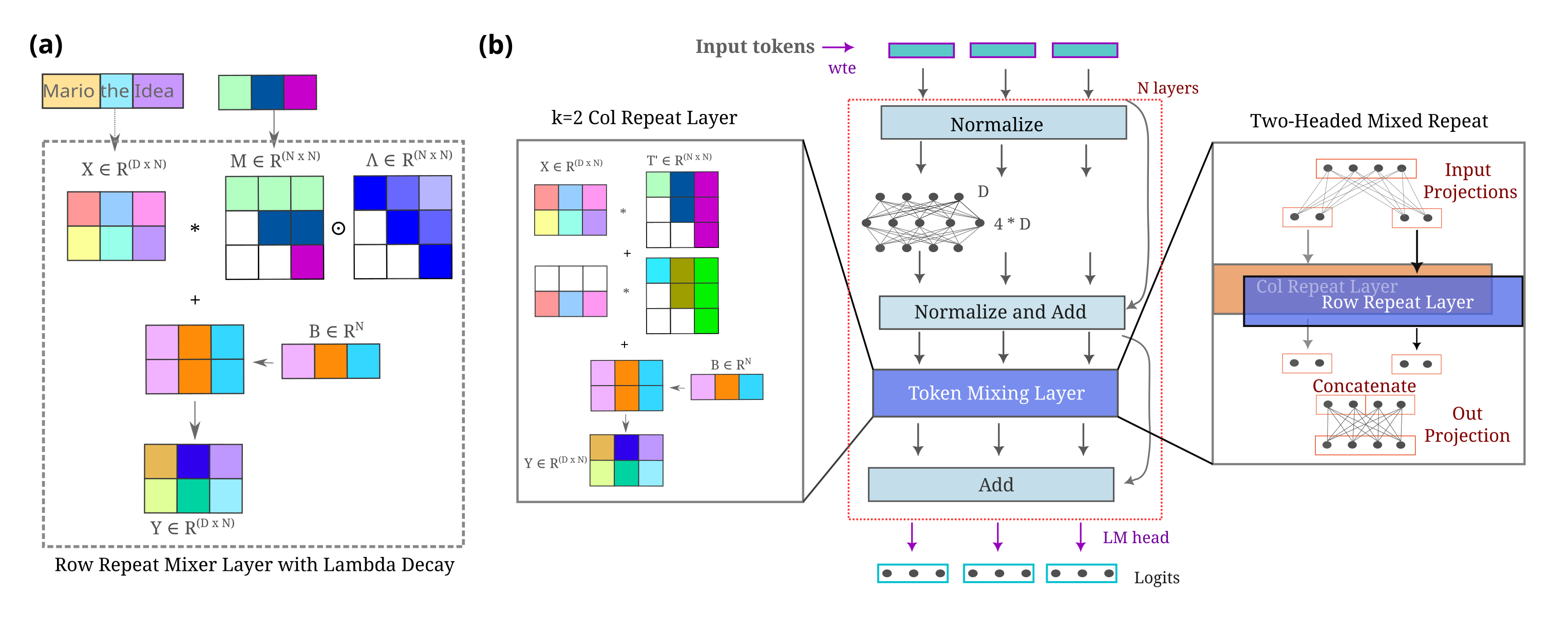}
        \caption{SRM architecture overview}
        \label{fig2}
    \end{figure}

    \begin{equation}
        Y = P_{out}\bigg ((I_0X)W_0 + \beta_0 \circ (I_1X)W_1 + \beta_1   \circ \cdots \circ I_{h-1}(X)W_{h-1} + \beta_{h-1} \bigg )
        \label{eq10}
    \end{equation}

    We also implement the SRM recurrent operation for non-unitary kernels, where each token mixing layer mixes along both sequence and a limited number of hidden dimension elements, with the number of hidden dimension elements mixed equal to the kernel size. We train separate filters for each kernel with weights $W \in \Bbb R^{k \times n \times n}$ parameterized as column repeat matrices (Equation \ref{eq3}). In our implementation, shown in Equation \ref{eq11}, we zero pad $X\in \Bbb R^{d + k \times n}$ and use a stride of size 1.

    \begin{equation}
    Y = \sum_{i=0}^k X_{(i:d+i, \;:)}\;W_i + \beta
    \label{eq11}
    \end{equation}

    Algebraic equivalence allows the SRM to be represented by either recurrent or sequence parallel operations without any change to the underlying cache structure or model parameters: for instance, we implement an SRM as a Pytorch module can perform either parallel or recurrent forward passes on-demand without modification to the underlying memory, which greatly simplifies pipelines in which fast, parallelized inference needs to be efficiently swapped with sequence parallel forward/backward passes for gradient formation and backpropegation (typical of online reinforcement learning).

    \subsection{Computational Complexity}
    
    The computational complexity of SRMs at inference, as for other recurrent models, is $\mathcal {O}(nd^2)$ time and $\mathcal{O}(d)$ space. It is notable that SRMs are `true' constant-space models in that contain constant factors which themselves do not scale with sequence length (see for example \citep{qin-zhong-2023-accelerating}).
    
    It is worth noting that the quadratic complexity of SRMs at train time is effectively hidden by the much larger constant factors associated with memory transfer and gradient production for token hidden layers, causing these models to resemble $nd^2$ as long as $n_{ctx}$ is not substantially larger than $d$. This is also true of Transformers, and accounts for the lack of increased throughput of models that train with linear complexity (Mamba for example) over SRMs at train time with increased context (Tables \ref{tables4} - \ref{tables7}).

    Global memory access on device (GPU) typically requires an order of magnitude more clock cycles than matrix arithmetic operations, and as such is limiting for inference throughput for Transformers. Batching inputs allows for more efficient memory access patterns, but still does not typically result in tensor core saturation for transformer models during inference. In the batch dimension $b$, for $n$ tokens the memory scaling characteristics of each transformer layer is $\mathcal{O}(nbd)$, whereas for each layer of the SRM the corresponding scaling is $\mathcal{O}(bd)$. 
    
\section{Experimental Results}

    \subsection{SRM architecture optimization}

    Experimentation shows that multi-headed SRMs are more efficient to causally train than non-headed models, that using both column and row repeat matrices outperforms row repeat alone or column repeat alone, and that models with a decay term are more efficient to train than SRMs without decay terms (Table \ref{tables0}).  The most efficiently trainable architecture we tested consisted of `mixed' heads such that half of the heads are row repeat and half column repeats, each with hidden dimension $d/h$ such that the total cache size per layer is $d$. We find that use of unique projections leads to slightly more efficient training than the use of parallel heads, and that curiously the use of in and out projections results in a decrease in per-step loss despite their redundancy, although there is an approximately 10\% training throughput cost to using projections compared to not using them,(Table \ref{tables0} - \ref{tables2}). We employ head projections in all models unless otherwise noted. Pytorch-based recurrent and sequence-parallel SRM mixer representation implementations are confirmed to be numerically stable and yield identical outputs even when cache values and decay constants are used in half precision.

    \subsection{Causal Training Efficiency}

    As one of the largest challenges to recurrent models is efficient and stable causal language model training, we measured the training efficiency of SRMs as compared to both quadratic-complexity and other linear-complexity models. For Transformers, we use the Llama 2 architecture \citep{touvron2023llama2openfoundation} featuring RoPE, RMS norms and improved parameter initialization which have been shown to outperform early causal Transformer architectures, and models of $d_m = x$ achieve similar training throughputs of Mixers with $d_m=2x$ over the range of $d_m$ observed here  \citep{badger2025maskedmixerslanguagegeneration}. We observe training efficiency through the lens of real world limitations, specifically GPU throughput and memory required, which can be converted to loss achieved per FLOP applied during training.

\begin{figure}[h]
        \centering
        \includegraphics[width=0.99\textwidth]{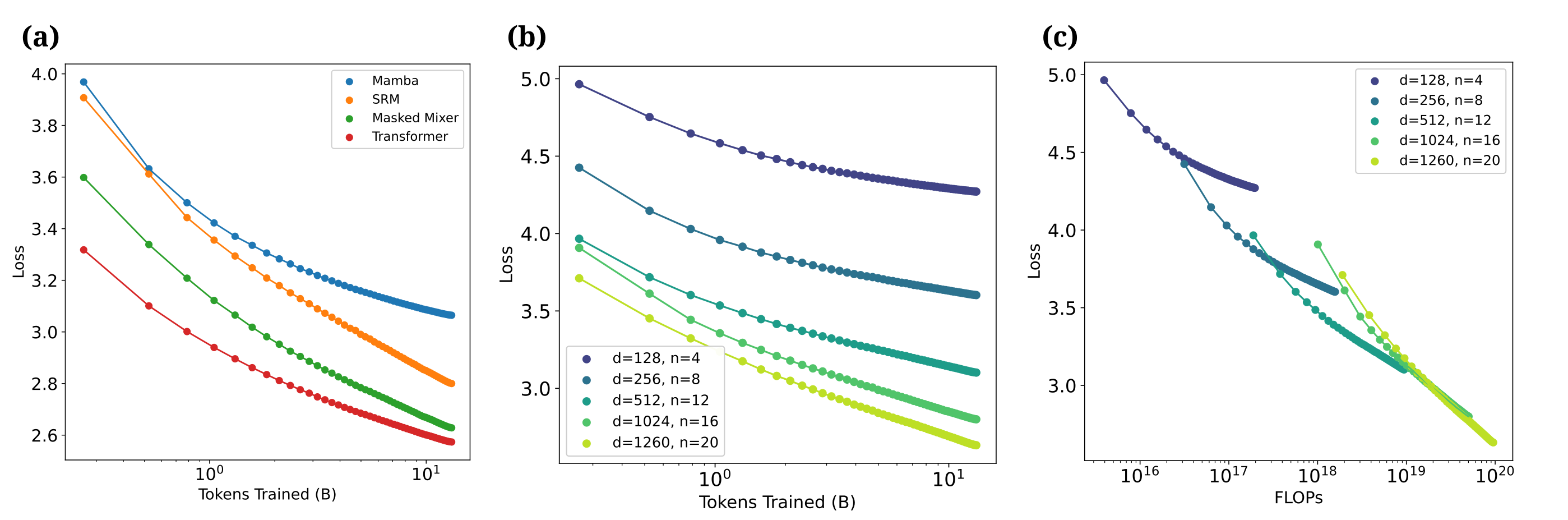}
        \caption{SRM causal training on FineWeb. (a) Throughput- and memory-equivalent model training: $d_m=1024$ SRM, $d_m=512$ Transformer, $d_m=256$ Mamba, all $n_l=16$. (b) SRM sample scaling characteristics for fixed model sizes, (c) SRM training compute (assuming theoretical maximum FLOPs on device as measured by V100 throughput) and loss scaling.}
        \label{fig3}
    \end{figure}

    We find that the SRM is slightly less efficient to train than $\mathcal O(n^2d^2)$ complexity Transformer and Masked Mixer models of similar sample throughput across diverse language understanding (FineWeb-edu) and mathematics and coding (FineMath) text corpora. For these datasets, we find that SRMs are more efficient to train than Mamba (2) models (Figure \ref{fig3}, Tables \ref{tables10}, \ref{tables11}). We note that we find Mamba 2 models (hereafter referred to as `Mamba' for simplicity as we do not investigate Mamba 1 due to that model's low training and inference throughput) to be only slightly less efficiently trainable than Transformers of similar throughput on a per-parameter basis (Table \ref{tables10}) as reported by \citep{dao2024transformersssmsgeneralizedmodels}, but that the substantially lower throughput on device (for H100s see Table \ref{tables1}, \ref{tables6}) for parameter-equivalent models results in more efficient SRM training in practice. On older devices (V100s) Mamba models experience a >20x drop in throughput and are virtually untrainable as a result, whereas SRMs perform consistently across all devices tested.
    
    Early in training RWKV (4) models exhibit similar efficiency as SRMs but suffer from catastrophic numerical instabilities across various hyperparameter settings, compute datatypes, and architectural configurations and as such are not explored in detail.

    \subsection{Benchmark Evaluations}

    To assess how the training efficiencies impact benchmark performance, we compare functional characteristics of linear-complexity models by evaluating them on a number of single-sample benchmark tasks without regard to the compute required to generate samples. We benchmark for general question answering via Arc-Easy \citep{Clark2018arc}, sentence completion using HellaSwag \citep{zellers2019hellaswagmachinereallyfinish} and Lambada (OpenAI processed) \citep{paperno2016lambadadatasetwordprediction}, information retention via SQuAD \citep{rajpurkar2016squad100000questionsmachine}, SQuAD 2 \citep{rajpurkar-etal-2018-know}, Longbench \citep{bai2024longbench}, IFEval \citep{zhou2023instructionfollowingevaluationlargelanguage}, SWDE \citep{arora2024swde}, and xWinoGrad \citep{tikhonov2021heads} using Eleuther-AI's evaluation harness \citep{eval-harness}. In Table \ref{table6} we detail benchmarks for compute-equivalent Transformer, SRM, and Mamba models, and we find that Transformers top the most benchmarks of all three models and that SRMs outperform Mamba (2) models on most benchmarks observed.

    \begin{table}
    \centering
    \footnotesize
    \renewcommand{\arraystretch}{1.1}
    \begin{tabular}{l c c c} 
      \textbf{Benchmark} & \textbf{Transformer} & \textbf{Mamba} & \textbf{SRM} \\
      \hline
      IFEval (strict) & 12.2 & \textbf{25.66} & 21.22 \\
      SQUAD (completion) & \textbf{22.96} & 10.22 & 1.47 \\
      SQUAD v2 & 5.1 & 1.69 & \textbf{16.87} \\
      LongBench (avg. acc) & 4.44 & \textbf{4.96} & 4.09 \\
      SWDE (contains) & \textbf{21.1} & 2.34 & 3.50 \\
      xWinoGrad & \textbf{57.7} & 51.52 & 53.41 \\
      GSM8k & \textbf{1.90\scriptsize±1.36} & 1.36\scriptsize±0.32 & 1.44\scriptsize±0.33 \\
      ARC (easy) & 48.11\scriptsize±1.03 & 33.96\scriptsize±0.97 & \textbf{50.0\scriptsize±1.03} \\
      WikiText (BPB, $\downarrow$) & 1.5192 & 1.6212 & \textbf{1.2363} \\
      GLUE (average acc) & 40.3 & 0.4552 & \textbf{0.4577} \\
      Lambada-OpenAI & \textbf{15.13\scriptsize±0.5} & 1.01\scriptsize±0.1& 3.05\scriptsize±0.24 \\
      HellaSwag & \textbf{29.6} & 26.74 & 28.86  \\
      \hline
    \end{tabular}
    \caption{Training compute-matched SRM, Transformer, and Mamba benchmark scores after FineWeb-edu training. All benchmarks are higher is better apart from WikiText.}
    \label{table6}
    \end{table}
    
    \subsection{Informational Retention and Capacity}

    The ability of quadratic-complexity models to retain input information in their hidden layers is of relatively minor importance for causal language modeling for the reason that these models are exposed to a cached representation of each input token upon each prediction step, and the prediction itself typically requires the knowledge of only a small subset of previous tokens. For linear-complexity models, however, the model is only exposed to a fixed-size cache of all previous tokens and cannot simply reference individual token representations separately. SRMs in particular hold a cache size equivalent to one token's embedding per layer for the entire input.  This motivates the measurement of the amount of information present in the hidden states of causal-trained models (which we term `information retention') as well as the ability of these models to store input information in one or more states (`information capacity', or equivalently the upper bound of information retention). These questions are especially important to SRMs because Masked Mixers trained for maximum information retention contain qualitatively different token mixing matrices than those of models trained for next token prediction (Figure \ref{fig1}, \ref{figs1}) and SRMs are by design only reflect those present for next token prediction.

    We measure model information retention by testing the capability of a CLM-trained model's embedding to regenerate input via a trained decoder. Given the output (i.e. the vector directly responsible for a single token prediction, namely the last hidden layer's last embedding) of token-predicting model $\mathcal{E}$ applied to input $x$, $y = \mathcal E(x, \theta_e)$, we seek to invert $y$ to recover $x$. We do this by first freezing $\theta_e$ and then training a decoder to recover $x$ given an unrolled form of $y$, where the decoder yields $z = \mathcal{D} \left( U(\mathcal{E} (x, \theta_e), \theta_d \right)$ and the difference in $x$ and $z$ in terms of both Cross-Entropy Loss as well as the more interpretable entropy ratio, the fraction of input information the model retains, after \citep{badger2026languagemodelmemorymemory}.

    We find that causal SRMs exhibit higher information retention than causal Mamba models, and that these differences exist before and after training (Table \ref{table3}); furthermore comparing our results to previous work \citep{badger2026languagemodelmemorymemory} we find that SRMs exhibit similar information retention characteristics to quadratic complexity models. We measure information capacity, or the amount of information a model is capable of retaining from its inputs, in two ways: encoder-decoder style input regeneration and copy task training efficiency. The encoder-decoder input regeneration is identical to that detailed above except the encoder $\theta_e$ is initialized from scratch and is unfrozen during training. The copy task is a measure of a model's ability to learn to copy 512 tokens in relatively few (10k instead of 200k) training steps. Copying may be thought of as a less information-intensive task than regeneration, such that copy task training illustrates a model's ability store smaller amounts of information relatively little training whereas input regeneration measures a models' ability to store more information with more training.
    
    We find that SRMs exhibit higher informational capacity as measured by encoder-decoder input regeneration (Table \ref{table4}) and similar or slightly better copy task training efficiency when adjusted for compute but not on an equivalent embedding size basis, with variation dependent on SRM architecture (Table \ref{table5}). To conclude, the qualitative features transferred to SRMs from causal Masked Mixers typically do not reduce information capacity, and SRMs exhibit superior information retention and information capacity to Mamba models on a per-compute basis.

     \begin{center}
    \begin{table}[H]
    \small
    \begin{center}
    \renewcommand{\arraystretch}{1.2}
    \begin{tabular}{|l c c c c |} 
    \hline
       & SRM & Untrained SRM & Mamba & Untrained Mamba \\
       \hline
       Loss $\downarrow$ & 6.169 & 6.105 & 7.039 & 7.290 \\
       Entropy Ratio $\uparrow$ & 0.3168 & 0.3239 & 0.2204 & 0.1926 \\
     \hline 
    \end{tabular}
    \end{center}
    \vspace{0.1cm}
    \caption{Causal pretrained model information retention. All models $d_m=512, n_l=16, n_{ctx}=512$}
    \label{table3}
    \end{table}
    \end{center}

    \begin{center}
    \begin{table}[H]
    \small
    \begin{center}
    \renewcommand{\arraystretch}{1.2}
    \begin{tabular}{|l c c c c c c |} 
    \hline
       & Mixed Heads & Decay & Parallel Heads & Head Projections & Loss & Entropy Ratio \\
       \hline
       SRM, H=4 & + & + & - & + & 2.011 & 0.7773 \\
       SRM, H=4 & + & - & - & + & 4.050 & 0.5514 \\
       SRM, H=4 & + & + & - & - & 1.295 & 0.8565 \\
       SRM, H=4 & - & - & - & + & 2.984 & 0.6695 \\
       SRM, H=4 & + & + & + & + & 2.423 & 0.7316 \\
       Mamba, H=8 & & & & & 3.357 & 0.6282 \\
       Masked Mixer &   &   &   &  & 0.440 & 0.9513 \\
     \hline 
    \end{tabular}
    \end{center}
    \vspace{0.1cm}
    \caption{Architecture Information Capacity, measured via encoder-decoder input reconstruction. All models $d_m=512, n_l=16, n_{ctx}=512$}
    \label{table4}
    \end{table}
    \end{center}

    \begin{center}
    \begin{table}[H]
    \small
    \begin{center}
    \renewcommand{\arraystretch}{1.2}
    \begin{tabular}{|l c c c c c c |} 
    \hline
       & Mixed Heads & Decay & Parallel Heads & Head Projections & Accuracy & Loss \\
       \hline
       SRM, d=1024 & + & + & - & - & 0.7474 & 0.9015 \\
       SRM, d=512 & + & + & - & - & 0.6886 & 1.201 \\
       SRM, d=256 & + & + & - & - & 0.3484 & 3.592 \\
       SRM, d=256 & - & - & - & + & 0.4043 & 3.084 \\
       SRM, d=256, H=0 &  &  &  &  & 0.442 & 2.802 \\
       SRM, d=256, k=4 &  &  &  &  & 0.4915 & 2.487 \\
       SRM, d=512, k=4 &  &  &  &  & 0.7854 & 0.8273 \\
       SRM, d=512, k=8 &  &  &  &  & 0.8439 & 0.5830 \\
       Mamba, d=128 & & & & & 0.252 & 4.46 \\
       Mamba, d=256 & & & & & 0.5121 & 2.437 \\
       Mamba, d=512 & & & & & 0.9673 & 0.133 \\
     \hline 
    \end{tabular}
    \end{center}
    \vspace{0.1cm}
    \caption{Copy accuracy (512 tokens copied at 10k training steps) $d_m=512, n_l=16, n_{ctx}=1024$. `k' denotes kernel number for SRM mixer layers, and H=0 denotes a non-headed model. Most SRM configurations are typically less efficient copy learners per $d_m$ size but more efficient per compute than Mamba models and achieve high copy accuracy with limited training steps.}
    \label{table5}
    \end{table}
    \end{center}

    These experiments suggest that while SRMs are capable of capturing nearly all of the information present in a reasonably sized input, causal-trained SRMs are typically not capable of doing so. This is true for other model architectures trained for causal modeling \citep{badger2026languagemodelmemorymemory}, but is particularly detrimental for recurrent models because they store fixed-sized caches. Recurrent models may be thought of as forming a compression of the input in their cache, and in order to retrieve this information accurately the compression must be of high fidelity and therefore scaling $n_{ctx}$ beyond the limit of compression achievable by the model necessarily leads to information loss and poor performance. Compression limits are highly dataset-dependent, and for low-information data such as DNA sequences are much lower (in terms of bits required per byte of input) than they are for high-information inputs like language. For language, assuming a 0.54 bits per byte (14.8x) compression achieved by recent LLMs \citep{deepseekai2025deepseekv3technicalreport} a $d_m=1024$ embedding would be expected to be able to losslessly compress up to $1024 \;params * 2 \frac{bytes}{param} * 0.5 \frac{tokens}{byte} * 14.8 = 15155$ tokens, although this is likely an overestimation given that it assumes that decoding the cache results in no further compression. We test the idea that scaling recurrent models' context beyond their compression limit necessarily results in performance degradation by observing copy fidelity as the number of copied tokens increases for a fixed training dataset and SRM model, and find a decline in copy accuracy (from >90\% to <30\%) as the number of copied tokens increases (Figure \ref{figs2}). This motivates the following investigation into scaling recurrent models in the batch, rather than the sequence, dimension.

    \subsection{Recurrent Throughput and Concurrency}

    For causal language model inference, tokens must be generated sequentially (or perhaps a few at a time) as the true distribution is not known \textit{a priori}, such that parallelization across the sequence dimension is no longer possible. Instead one can parallelize across the batch, or sample, dimension with the relatively safe assumption that the samples are independent. It is clear that the recurrent representation of SRMs is more efficient to parallelize in the sample dimension than linear space models such as the Transformer, as for $s$ samples and $l$ layers a recurrent model requires $sld$ space per layer and at each next token prediction transfers $sld$ elements from memory, whereas Transformers require $sldn_{ctx}$ space and transfer $sldn_{ctx}/2$ elements from memory on average. If memory transfer alone was limiting, this means that one would expect to observe a throughput increase of $(n_{ctx}/2)$x in SRMs compared to Transformers of equivalent size. Asymptotically as $n_{ctx} \to \infty$ this would be true of real devices, but for our context lengths the penalty for larger batch sizes through the token feedforward layers is non-trivial. Even for the relatively small context length sizes used in this work, however, we find that there is an order-of-magnitude increase in throughput (Table \ref{table1}). Likewise, we would expect for an $n_{ctx}$x increase in the memory required in Transformers compared to SRMs as $n_{ctx} \to \infty$, and we observe a two orders-of-magnitude increase in maximum concurrency (Table \ref{table1}). 

    \begin{center}
    \begin{table}[H]
    \small
    \begin{center}
    \renewcommand{\arraystretch}{1.2}
    \begin{tabular}{|l c c | c c | c |} 
    \hline
      & \multicolumn{2}{c}{Transformer} & \multicolumn{2}{c}{SRM} & \\
      $n_{ctx}$ & Throughput (t/s) & Max Concurrency & Throughput (t/s) & Max Concurrency & Throughput Increase \\
     \hline \hline
      512 & 2908 & 400 & 28091 & 64000 & 9.66x \\
      1024 & 1918 & 200 & 28298 & 64000 & 14.75x  \\
      2048 & 1116 & 100 & 27441 & 32000 & 24.59x \\
      4096 & 634 & 50 & 27445 & 32000 & 43.29x \\
      \hline 
    \end{tabular}
    \end{center}
    \vspace{0.1cm}
    \caption{Throughput (tokens per second) per context length, 1x V100 (16GB) with Pytorch implementations, SRMs are $d_m=1024$ and Transformers are $d_m=512$, both $n_l=16, h=4$. Note that the decrease in batch size for $n_{ctx}>1024$ for SRMs is not due to the model using more memory, but due to the memory required to store generated tokens via a non-optimized in-memory store.}
    \label{table1}
    \end{table}
    \end{center}

    When we compare throughput and concurrency of the SRM to Mamba models, we find that the former experiences much higher ($>500x$) concurrency and approximately $7x$ the throughput of equivalently powerful models of the latter, although Mamba models are in turn observed to exhibit increased throughput compared to Transformers (Table \ref{table2}). We attribute this substantial difference to tiny number of transformations SRMs perform during sequence mixing (two vector-scalar multiplications and a vector addition per layer).

    \begin{center}
    \begin{table}[H]
    \small
    \begin{center}
    \renewcommand{\arraystretch}{1.2}
    \begin{tabular}{|l c c c c c c c|} 
    \hline
       & SRM$_{1024}$ & Mamba$_{256}$ & Mamba$_{512}$ & Transformer$_{256}$ & Transformer$_{512}$ & RWKV$_{256}$ & RWKV$_{512}$ \\
       \hline
       Throughput & 161312 & 61465 & 23155 & 32134 & 15272 & 3820 & 3774 \\
       Concurrency & 512000 & 1000 & 1000 & 6000 & 4000 & 1024 & 1024 \\
     \hline 
    \end{tabular}
    \end{center}
    \vspace{0.1cm}
    \caption{Maximum throughput (tokens per second) and concurrency (samples) per model, 1x H100 (96GB) with Pytorch implementations, aside from Mamba which uses optimized Triton device kernels (this architecture has much smaller throughput without the use of these kernels). All models are $n_{ctx}=512, n_l=8$, and RWKV is limited by default to 1024 batch size, and the maximum predicted batch for the $d_m=512$ model is around 2000.}
    \label{table2}
    \end{table}
    \end{center}

    To demonstrate the benefits of building a production-grade inference framework for recurrent models, we implemented recurrent SRMs using Modular MAX inference framework \citep{modularMAXHighperformance} and verified that implementing an SRM in MAX resulted in substantially higher throughput (7.5x) and concurrency (1.75x) compared to that obtained by Torch implementations using identical hardware. We then compared the throughput of SRMs to those of equivalently capable transformers applied using vLLM, a widely applied efficient inference engine built around paged attention \citep{kwon2023efficient}. We observe similar efficiency increases in our MAX SRM implementation compared to a Transformer inferenced via vLLM as we did for Pytorch implementations of both models: the SRM exhibits 12x the throughput and 170x the concurrency of a pretraining-matched (26x the throughput and 225x the concurrency of a parameter-matched) Transformer when generating 1k tokens on a single Nvidia H100 (Tables \ref{table11}, \ref{tables20}). On the optimized Albatross inference engine, the RWKV exhibits increased throughput compared to the Transformer but still exhibits only a fraction of the throughput of the SRM.
    
    \begin{center}
    \begin{table}[H]
    \small
    \begin{center}
    \renewcommand{\arraystretch}{1.2}
    \begin{tabular}{|l c c c c c |} 
    \hline
       & SRM$_{1024}$ & SRM$_{512}$ & Mamba & Transformer & RWKV \\
       \hline
       Throughput (k t/s) & 1203 & 2626 & 44* & 101 & 184 \\
       Concurrency (k samples) & 800 & 1200 & 20* & 5.32 & 50 \\
     \hline 
    \end{tabular}
    \end{center}
    \vspace{0.1cm}
    \caption{Maximum throughput and concurrency per model, 1x H100 (96GB) with optimized engine inferences (SRMs via our Mojo/MAX implementation, Transformers and Mamba on vLLM with optimized kernels if appropriate, RWKV (7) on Albatross). Non-SRM models are all $d_m=512$, and all are $n_l=8$ with an 8k-size tokenizer. Mamba models exhibit a memory leak and do not reach their theoretical limit of 20k concurrent samples, achieving only 300.}
    \label{table11}
    \end{table}
    \end{center}

    At the throughputs achieved by SRMs, the operation of finding the next tokens given the model's logits becomes a bottleneck: for example, the throughput of the $d_m=512$ with argmax on the logits to compute the next token is only 63\% that of the same model without applying argmax (Table \ref{tables20}).

    We comment that these results illustrate the versatility of the SRM architecture: unlike Mamba and RWKV which specialized device kernels for performant inference, SRMs are easily incorporated into other inference frameworks and do not require substantial rewrites for new hardware. This is a significant problem for training as well as inference: Mamba kernels perform poorly on older hardware (V100s) due to differences in memory management, and both Mamba and RWKV kernels will likely need to be rewritten for new GPU architectures in order to maintain performance as L1/L2/shared cache memory management is likely to change in the future.

    \subsection{Verifiable Output Scaling}

\begin{figure}[h]
        \centering
        \includegraphics[width=0.9\textwidth]{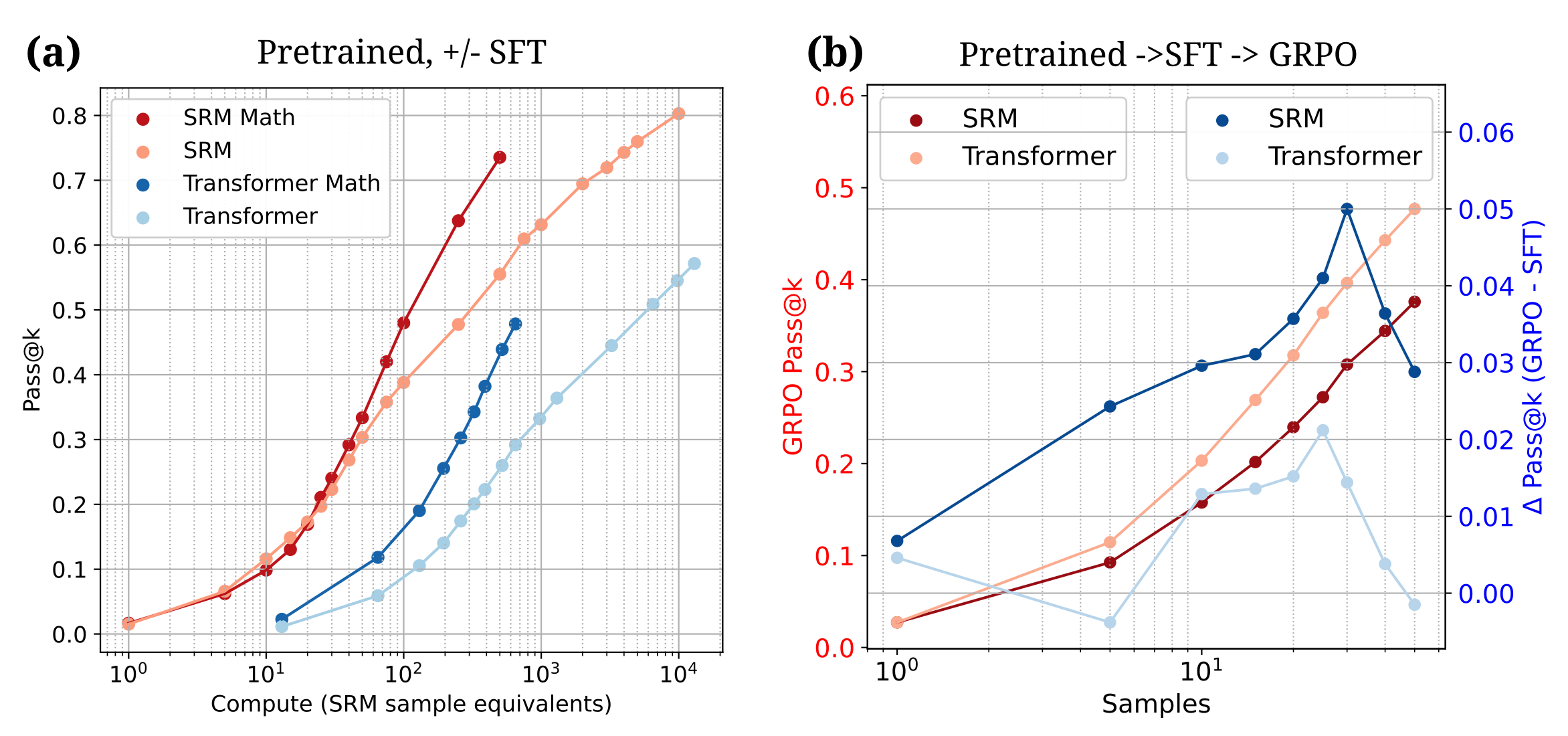}
        \caption{(a) GSM8k Inference Scaling, in compute per SRM sample. `Math' models are pretrained on FineMath 4+ followed by SFT on GSM8k, others are pretrained only on FineWeb. (b) Compute-constant GRPO (50 samples per batch for SRM, 5 for Transformers) on Math-trained models, with balanced resampling for SRMs. SRMs are $d_m=1024$ and Transformer $d_m=512$}
        \label{fig4}
    \end{figure}

\begin{figure}[h]
        \centering
        \includegraphics[width=0.99\textwidth]{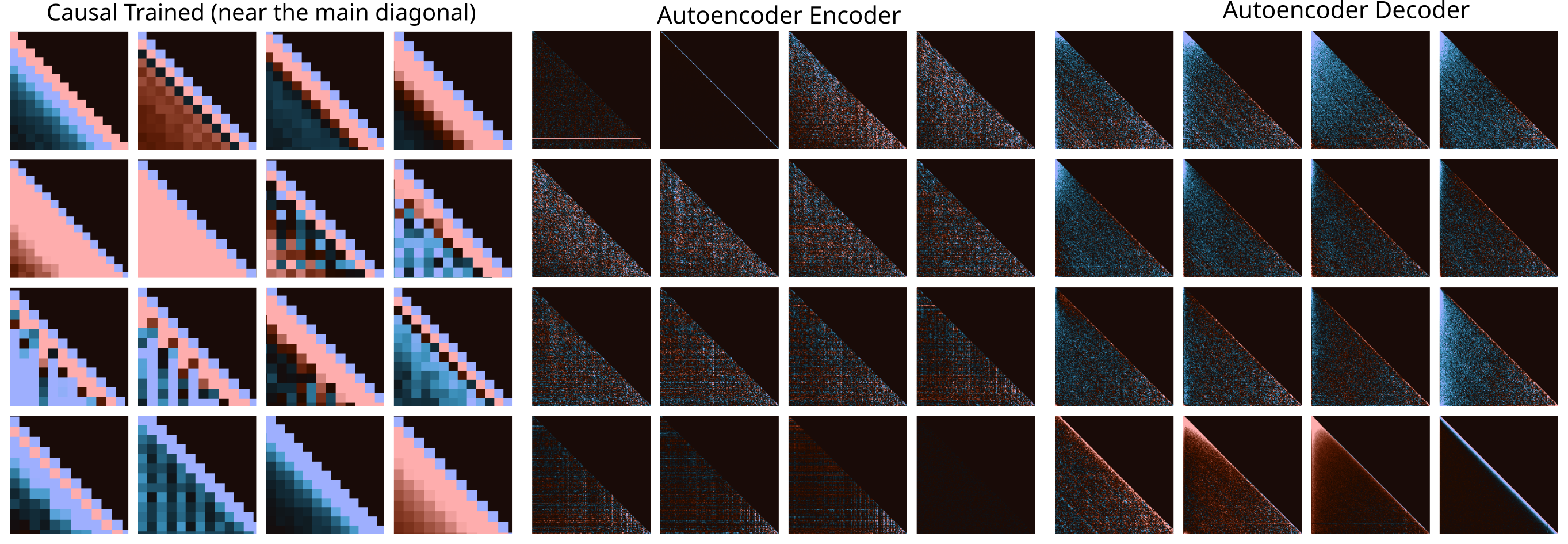}
        \caption{Token mixing weights from trained masked mixers. Autoencoders trained for maximum information retention (into the last hidden layer of the last token).}
        \label{figs1}
    \end{figure}

\begin{figure}[h]
        \centering
        \includegraphics[width=0.4\textwidth]{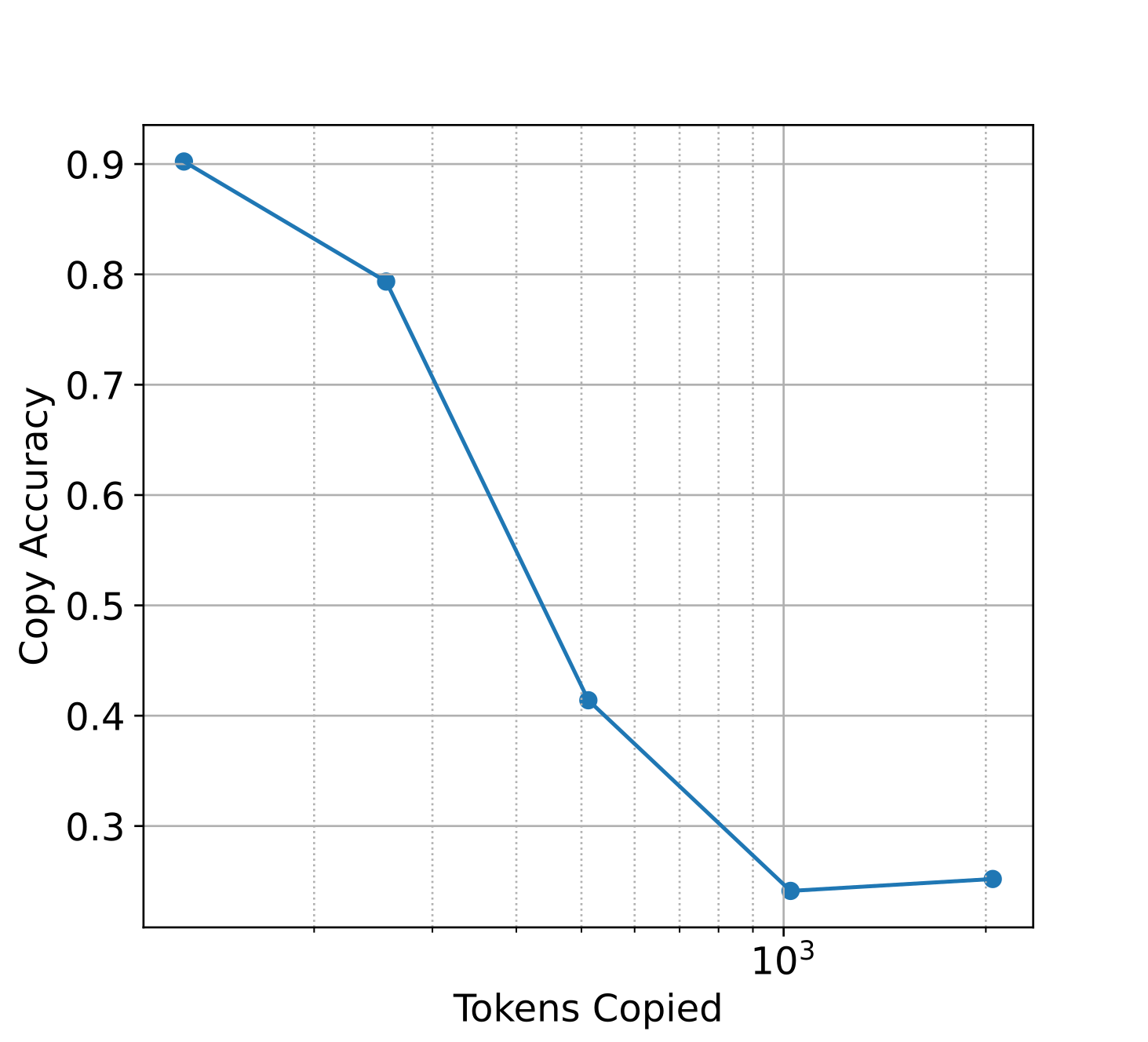}
        \caption{Copy accuracy plummets as the number of tokens to be copied increases. SRM, $d_m=256, n_l=16$, the total context window is double the tokens copied (FineWeb corpus), trained for 10k steps, $b=64$ samples per step.}
        \label{figs2}
    \end{figure}

    We next investigated whether the 10-30x throughput increase for SRMs versus Transformers results in materially higher performance on benchmarks in which outputs may be verified, and in particular we focus on GSM8k with verification in the form of an oracle that tests equivalence of the model's last numerical output for that question with the correct answer. The following question is addressed: given a fixed amount of compute and memory (over both train and test time), what is most efficient for obtaining at least one correct sample per question?  Equivalently, how does Pass@k scale with increased $k$ for SRMs, and in particular does it follow a similar power law found for Transformers across many tasks \citep{brown2024largelanguagemonkeysscaling}? We find that the Pass@k scaling for SRMs is nearly identical to that for Transformers for GSM8k (Figure \ref{fig4}). In practical terms, this means that to reach a given target coverage the transformer requires around 10x the inference compute, and for a constant compute budget the SRM gives a correct answer to around 30\% more GSM8k questions than the Transformer. We find that this relationship holds regardless of preparation for the GSM8k dataset: samples from SRMs and Transformers pretrained on 13B tokens of FineMath-4+ followed by SFT on the GSM8k training dataset are by eye more coherent and exhibit higher Pass@k accuracy than those from models trained on FineWeb, but again SRMs exhibit around 30\% more coverage per fixed compute.

    \subsection{Reinforcement Learning}

    Greater inference throughput afforded via SRMs would be expected to benefit reinforcement learning methods that make use of parallel generation with the goal of maximizing the probability that one or a few outputs are correct. One class of this criteria is where a model is used to generate multiple token sequences (rollouts) independently in parallel, which are then evaluated and reinforced depending on whether they fulfill some criteria. This class encompasses PPO \citep{schulman2017proximalpolicyoptimizationalgorithms} as well as GRPO \citep{shao2024deepseekmathpushinglimitsmathematical} and its variants \citep{liu2025understandingr1zeroliketrainingcritical}. With some simplification, GRPO proceeds by first generating a group $G$ of samples from a model for a given question, verifying the goodness of each sample using reward functions, and then performing gradient descent to change the model's parameters such that `good' outputs with higher rewards are more likely to be sampled whereas bad outputs are less likely. Parallel generation coupled with random sampling (typically top-p nucleus sampling \citep{holtzman2020curiouscaseneuraltext} with temperature) results in distinct samples usually containing non-identical token sequences. In this work we evaluate on outcomes only, such that the reward $r=1.0$ if evaluation is correct and $r=0.0$ if incorrect.

    \begin{equation}
    \mathcal{J}= \Bbb E[q \sim P_q, \{G_r\} \sim \pi_{\theta} (\{o_i \}_{i=1}^{G_r}|q)] \frac{1}{G_r} \sum_i \frac{1}{|o_i|} \sum_{t=1}^{|o_i|} \left( A_{i, t} - \beta \mathbb{D}_{KL}(\pi_{\theta}|| \pi_{\theta_{ref}}) \right)
    \label{eq9}
    \end{equation}
    
\begin{figure}[h]
        \centering
        \includegraphics[width=0.85\textwidth]{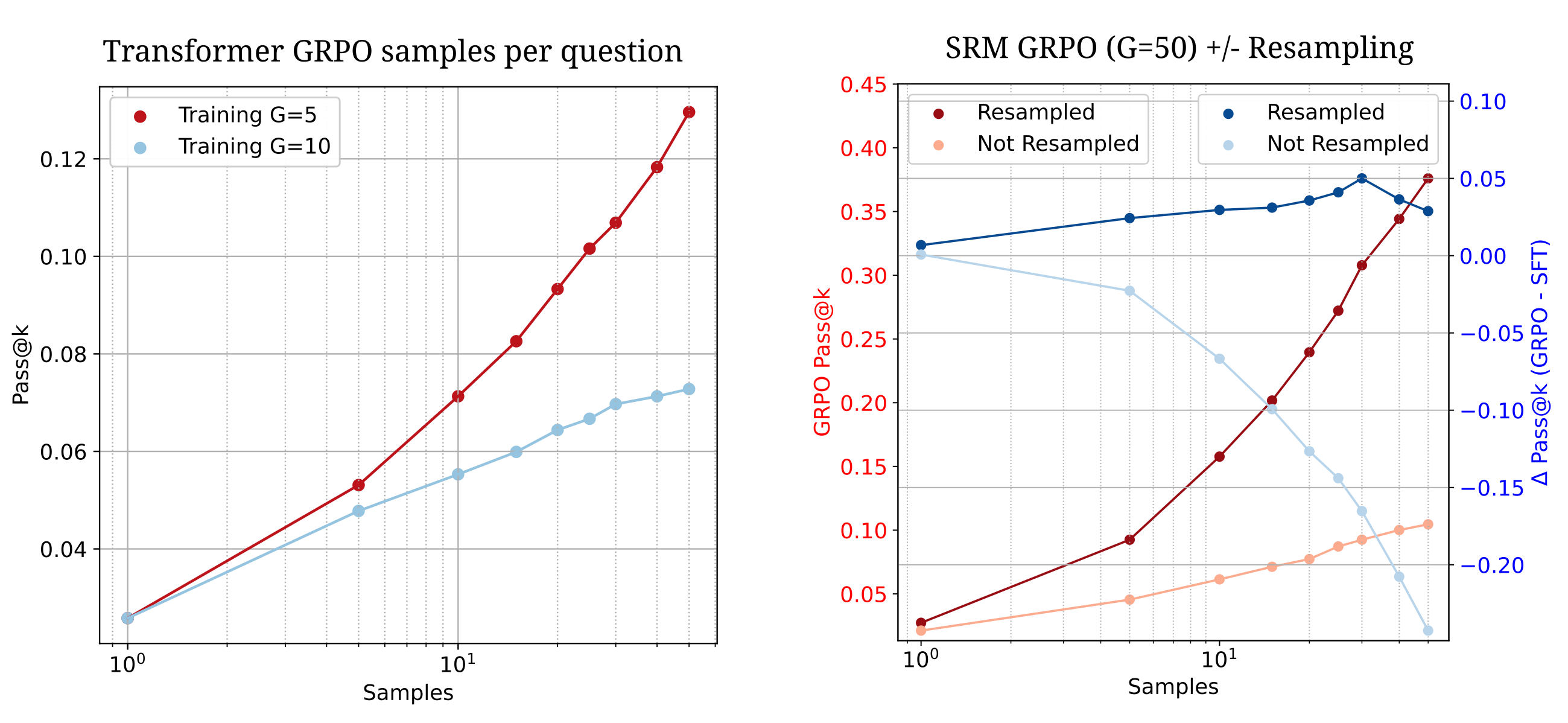}
        \caption{GRPO exploration on GSM8k. Models are pretrained on 13B tokens of FineMath 4+ followed by SFT on GSM8k training dataset. }
        \label{figs3}
    \end{figure}

    One of the primary challenges with parallelized rollout RL algorithms such as GRPO is the inherent limitation that the model must be capable of generating at least one `good' output of the batch (here where $r=1.0$), or the model has no reference with which to better approximate. GRPO is particularly susceptible to decreases in model exploration when few samples are rewarded because an implicit penalty is applied for each incorrect sample which limits exploration relative to SFT-trained models, which manifests in lower Pass@k (k>1) accuracy of GRPO- versus SFT-trained models \citep{shao2024deepseekmathpushinglimitsmathematical}, which we find worsens with as the generation batch size $G$ increases (Figure \ref{figs3}). To prevent this, we introduce a balanced resampling approach where for each sampled batch $G$ of generations per question, given a training batch size $b$ the generations are evaluated and we resample $G$ to include up to $b/2$ `good' samples, before resampling $G$ selecting for `bad' samples to make the training batch reach size $b$. Denoting this resampled $G$ as $G_r$, and simplifying for the case where one training step is taken per generation batch such that $\pi_{\theta_{old}} = \pi_{\theta}$ then consequently $\pi_{\theta}(o_{i}|q, o_{i}) / \pi_{\theta_{old}}(o_{i}|q, o_{i}) = 1$, and thus we maximize the objective function given in Equation \ref{eq9} when batches are large (ie for SRMs) and use the standard GRPO objective for small-batch training. In practice we typically pool samples from multiple questions per $G$, such that resampling occurs over these pooled batches.
    
   With FineMath 4+ pretrained models, we performed SFT on GSM8k followed by GRPO and confirmed that balanced resampling substantially increases Pass@k for SRMs (Figure \ref{figs3}).  We then compared GRPO training of SRMs generating 50 samples per question versus Transformers with 5 samples per question, both with a single forward/reverse pass per batch and SRMs using balanced resampling. The results of both Pass@k after GRPO, as well as the difference between models before and after GRPO are shown in Figure \ref{fig4} (b), and we observe that the single-sample accuracy is slightly higher for SRMs than for Transformers. The tradeoff between higher single-sample accuracy and lower exploration is clearly visible for Transformers, but in contrast we find that SRMs with their larger, resampled batches lead to increased successful exploration.

\section{Discussion}

    \subsection{Limitations of this work}

    The SRM architecture exhibit a few notable limitations: firstly as implemented it is not suitable for modeling samples where the sequence dimension far exceeds the hidden dimension due to the quadratic complexity (w.r.t. token length) at training, secondly there is no built-in method for context length and candidate methods have yet to be tested, and finally the architecture is most easily optimized for larger parameter numbers with fewer activations compared to Mamba or Transformer models, and is unsuitable for applications requiring minimal parameter counts. This work is limited in scale and also in scope, and does not explore a number of optimizations (parallel scan for recurrent prefill, for example). One of the main outstanding question relates to the benefits or drawbacks one receives for various classes of problems when using a Transformer compared to an SRM: is it better to use a somewhat more capable model yielding fewer samples or a somewhat less capable one that gives >10x the throughput?

    \subsection{The Future}

    Many-threaded versus single-threaded performance has increased exponentially the last couple of decades, as has the ratio of global memory bandwidth to arithmetic operation latency (the `memory wall') on device \citep{kirk2016programming, gholami2024aimemorywall}, and it is likely that these features will continue for the foreseeable future. Although there is a limit to how parallelizable current inference scaling techniques such as Chain of Thought \citep{wei2023chainofthoughtpromptingelicitsreasoning} are, with some caveats one would expect for the performance (i.e. useful sample production efficiency) of parallel sample generation to far outstrip single-sample sequential generation in the years to come because of the correspondence of parallel sample generation to parallel threads on device. Of the architectures studied in this work, SRMs are the most suitable for parallel generation and furthermore are ideally situated for dealing with the memory wall, as they exhibit a higher ratio of arithmetic operations to memory accesses. If these thread and memory characteristics hold in the future, SRMs will continue to increase their relative efficiency compared to memory-intensive models such as Transformers. Signs of practical aspects of this can be seen in this work: the ratio of SRM to Transformer throughput on V100s at $n_{ctx}=512$ is 9.66x, whereas for the newer H100 is 10.56x (using compiled Pytorch implementations).

\section{Acknowledgments}

    The author is grateful to Collin Stedman for many helpful discussions on the nature of modeling bottlenecks and reinforcement learning tradeoffs, and to Matthew Neligeorge for conversations on the utility of recurrent models from an efficiency standpoint.

\bibliographystyle{unsrtnat}
\bibliography{references}  

\beginsupplement
\section{Appendix}

    \subsection{Training Details}

    Pretraining proceed as follows: first datasets were pretokenized and padded as necessary, and models were initialized and compiled where possible and optimized via the Pytorch implementation of AdamW with $\beta_1=0.9, \beta_2=0.999$. Most pretraining datasets use a constant token number of $200,000 * 512 * 128=13.1B$ tokens, and this is held constant for longer-context trainings (here we have $200,000 * 1024 * 64=13.1B$ tokens). We use $\eta=5*10^{-4}$ for SRMs and Mamba models and $\eta=2*10^{-4}$ for Transformers after optimizing this hyperparameter independently, and use a 4000-step warm-up with linear learning rate decay.
    
    We train for supervised finetuning by masking input tokens and training only on model-generated outputs, minimizing Cross-Entropy Loss between these and the answers provided by the GSM8k training dataset. We train with $\eta=1* 10^{-4}$, gradient clipping with maximum grad norm of 0.1, weight decay of 0.1, warmup ratio of 0.1, and optimize using the Pytorch implementation of AdamW with $\beta_1=0.9, \beta_2=0.999$, and selected the checkpoint corresponding to the minimum evaluation loss (which invariably corresponded to around the third epoch with these hyperparameters regardless of the model being trained). 

    GRPO training was performed using the `trl` \citep{vonwerra2020trl} GRPO trainer, and was modified for resampling and SRM-specific cache handling for that model. For all models we use $\eta=2*10^{-5}$, $\beta=0.04$, gradient clipping with a maximum grad norm of $0.1$, warm-up ratio of 0.1, weight decay of 0.1, and top-p sampling (top-p of 0.9, temperature=0.7) and optimize via the Pytorch implementation of AdamW.

    For all training runs, Pytorch models were compiled when possible and trained via DDP with automated fp16/fp32 mixed precision, except for Mamba autoencoder models which were trained via bf16/fp32 mixed precision training due to numerical stability issues. The main exceptions to compilation were SRMs with GRPO, where turning on and off the cache for switching between recurrent generation and sequence parallel training causes shape changes that are incompatible with older GPUs (V100s) although not with newer ones, and Mamba models for which we used optimized kernels that are themselves compiled and cannot be trivially re-compiled into the Pytorch graph.

    \subsection{MAX model details}

    MAX implementations are designed for demonstration only, and use a hybrid graph approach to maintain consistency with Pytorch implementations: the throughputs measured are for implementations of SRM Mojo/MAX static graphs compiled with with embedded dynamic Pytorch tensors for the cache vectors (we provide but did not profile purely static MAX graph implementations due to hardware constraints).

    When we profiled MAX SRMs, we found that there existed a significant (around 200ms) delay while calling empty MAX graphs from python driver code, and that this delay persists when calling SRM implementations with our driver code. Observing that forward passes and cache updates may be internally looped inside the forward pass code and thereby avoiding calling the graph multiple times per generation (although notably MAX does not currently support compilation of default argmax operations), we adjust for this delay in Table \ref{table11}, and provide the raw throughput values in Table \ref{tables20}.

    \begin{center}
    \begin{table}[H]
    \small
    \begin{center}
    \renewcommand{\arraystretch}{1.2}
    \begin{tabular}{|l c c c c|} 
    \hline
       & SRM$_{1024}$ & SRM$_{1024}$ +argmax & SRM$_{512}$ & SRM$_{512}$ +argmax \\
       \hline
       Throughput (k t/s) & 1231 & 900 & 2801 & 1764 \\
       Concurrency (k samples) & 800 & 800 & 1200 & 1200 \\
     \hline 
    \end{tabular}
    \end{center}
    \vspace{0.1cm}
    \caption{Raw H100 (NVL) MAX SRM throughputs, with delay from driver code calls.}
    \label{tables20}
    \end{table}
    \end{center}

    We found a bug in the current (26.1 and 26.2) MAX open source repository all machines tested: there exists a memory leak in which LLVM exhibits an out-of-memory error that depends on the number of Mojo modules called during model instantiation before compiling, without regard to the tensor size associated with those modules or what modules are called. For architectural consistency with Pytorch-profiled models (meaning that SRMs use four heads per layer, mixed row/column heads, nonparallel heads and input and output projections) this necessitated profiling models with fewer layers, and we chose 8 rather than 16. As the memory error is module call-dependent, we found that parallelized four-headed SRMs without input/output projections were able to be initialized with all 16 layers.

    \subsection{Benchmark Details}

    Functional benchmarks were performed using modified versions of Eleuther AI's LM evaluation harness \citep{eval-harness}. Code relating to SRM cache instantiation, updates, and destruction interfacing was added to a fork of the evaluation harness, and is available at https://github.com/blbadger/lm-evaluation-harness. Mamba and Transformer models used Huggingface implementations in that module. All benchmarks are zero-shot unless otherwise noted.

     \begin{center}
    \begin{table}[H]
    \small
    \begin{center}
    \renewcommand{\arraystretch}{1.2}
    \begin{tabular}{|l c c|} 
    \hline
      Model & Head Number &  Eval Loss \\
      \hline
      Row Repeat & 0 & 3.465 \\
      Column Repeat & 0 & 3.357 \\
      Row Repeat & 4 & 3.311 \\
      Column Repeat & 4 & 3.275 \\
      Combined Repeat & 4 & 3.219 \\
      Mixed Repeat & 4 & 3.228 \\
      Mixed Repeat +decay & 4 & 3.002 \\
      Combined Repeat +decay & 4 & 3.033 \\
      Mixed Repeat, +decay, -head projections & 4 & 3.056 \\
      \hline 
    \end{tabular}
    \end{center}
    \vspace{0.1cm}
    \caption{SRM training loss at 200k steps (13.1 B tokens), FineWeb-edu. All models $d_m=512, n_l=16$.}
    \label{tables0}
    \end{table}
    \end{center}
    
     \begin{center}
    \begin{table}[H]
    \small
    \begin{center}
    \renewcommand{\arraystretch}{1.2}
    \begin{tabular}{|l c|} 
    \hline
      Model & Eval Loss \\
      \hline
      Mamba $d_m=128$ & 3.41975 \\
      Mamba $d_m=256$ & 3.0647 \\
      Mamba $d_m=512$ & 2.758 \\
      SRM $d_m=512$ & 3.002 \\
      SRM $d_m=1024$ & 2.80 \\
      Masked Mixer $d_m=1024$ & 2.628 \\
      Transformer $d_m=512$ & 2.573 \\
      \hline 
    \end{tabular}
    \end{center}
    \vspace{0.1cm}
    \caption{FineWeb-edu training, 13B tokens, $n_{ctx}=512$. SRMs are mixed headed with decay.}
    \label{tables10}
    \end{table}
    \end{center}

     \begin{center}
    \begin{table}[H]
    \small
    \begin{center}
    \renewcommand{\arraystretch}{1.2}
    \begin{tabular}{|l c|} 
    \hline
      Model & Eval Loss \\
      \hline
      Mamba & 2.989 \\
      SRM (decay, mixed heads) & 1.614 \\
      Masked Mixer & 1.483 \\
      Transformer & 1.391 \\
      \hline 
    \end{tabular}
    \end{center}
    \vspace{0.1cm}
    \caption{FineMath 4+ training, 13B tokens, compute-equivalent}
    \label{tables11}
    \end{table}
    \end{center}

     \begin{center}
    \begin{table}[H]
    \small
    \begin{center}
    \renewcommand{\arraystretch}{1.2}
    \begin{tabular}{|l c|} 
    \hline
      Model & Eval Loss \\
      \hline
      +diagonal constant, - decay & 3.243 \\
      +diagonal constant, + decay & 2.981 \\
      -diagonal constant, - decay & 3.228 \\
      -diagonal constant, + decay & 3.002 \\
      \hline 
    \end{tabular}
    \end{center}
    \vspace{0.1cm}
    \caption{Constant diagonal models offer no consistent training efficiency increase. All $d_m=512, n_l=16$}
    \label{tables12}
    \end{table}
    \end{center}

    \begin{center}
    \begin{table}[H]
    \small
    \begin{center}
    \renewcommand{\arraystretch}{1.2}
    \begin{tabular}{|l c c c|} 
    \hline
      $d_m$ & Throughput & Memory & Parameter Count (M) \\
     \hline \hline
      128 & 17.01 & 7803 & 4.76 \\
      256 & 9093 & 9093 & 14.69 \\
      512 & 12.58 & 14663 & 50.29 \\
      1024 & 6.15 & 25257 & 184.40 \\
      2048 & 1.99 & 53463 & 704.28 \\
      \hline 
    \end{tabular}
    \end{center}
    \vspace{0.1cm}
    \caption{Training Throughput (samples/sec) and Memory (GB per device) and parameter count for SRM models ($h=4$, $n_l=16$, mixed heads, decay, nonparallel) on 2x H100 GPUs (NVL).}
    \label{tables1}
    \end{table}
    \end{center}

    \begin{center}
    \begin{table}[H]
    \small
    \begin{center}
    \renewcommand{\arraystretch}{1.2}
    \begin{tabular}{|l c c c|} 
    \hline
      $d_m$ & Throughput & Memory & Parameter Count (M) \\
     \hline \hline
      128 & 22.46 & 8839 & 4.76 \\
      256 & 19.73 & 12111 & 14.69 \\
      512 & 13.49 & 17247 & 50.29 \\
      1024 & 6.39 & 27093 & 184.40 \\
      2048 & 2.03 & 52431 & 704.28 \\
      \hline 
    \end{tabular}
    \end{center}
    \vspace{0.1cm}
    \caption{Training Throughput (samples/sec) and Memory (GB per device) for head-parallel SRM models (h=4, mixed heads, decay) on 2x H100s.}
    \label{tables2}
    \end{table}
    \end{center}

    \begin{center}
    \begin{table}[H]
    \small
    \begin{center}
    \renewcommand{\arraystretch}{1.2}
    \begin{tabular}{|l c c c|} 
    \hline
      $d_m$ & Throughput & Memory & Parameter Count (M) \\
     \hline \hline
      128 & 22.70 & 9571 & 4.23 \\
      256 & 19.33 & 11561 & 12.59 \\
      512 & 13.20 & 15703 & 41.89 \\
      1024 & 6.67 & 23959 & 150.80 \\
      2048 & 2.26 & 45373 & 570.00 \\
      \hline 
    \end{tabular}
    \end{center}
    \vspace{0.1cm}
    \caption{Training Throughput (samples/sec) and Memory (GB per device) for parallelized, non-projected SRM models (h=4, mixed heads, decay) on 2x H100s.}
    \label{tables3}
    \end{table}
    \end{center}

    \begin{center}
    \begin{table}[H]
    \small
    \begin{center}
    \renewcommand{\arraystretch}{1.2}
    \begin{tabular}{|l c c c|} 
    \hline
      $d_m$ & Throughput & Memory & Parameter Count (M) \\
     \hline \hline
      128 & 22.46 & 8839 & 4.76 \\
      256 & 19.73 & 12111 & 14.69 \\
      512 & 13.49 & 17247 & 50.29 \\
      1024 & 6.39 & 27093 & 184.40 \\
      2048 & 2.03 & 52431 & 704.28 \\
      \hline 
    \end{tabular}
    \end{center}
    \vspace{0.1cm}
    \caption{Training Throughput (samples/sec) and Memory (GB per device) for head-parallel SRM models ($h=4, n_{ctx}=512, b=128$, mixed heads, decay) on 2x H100.}
    \label{tables4}
    \end{table}
    \end{center}

    \begin{center}
    \begin{table}[H]
    \small
    \begin{center}
    \renewcommand{\arraystretch}{1.2}
    \begin{tabular}{|l c c c|} 
    \hline
      $d_m$ & Throughput & Memory & Parameter Count (M) \\
     \hline \hline
      128 & 17.29 & 7297 & 4.82 \\
      256 & 16.82 & 9603 & 14.76 \\
      512 & 12.23 & 14951 & 50.36 \\
      1024 & 6.17 & 25833 & 184.47 \\
      2048 & 2.05 & 53327 & 704.35 \\
      \hline 
    \end{tabular}
    \end{center}
    \vspace{0.1cm}
    \caption{Training Throughput (samples/sec) and Memory (GB per device) for head-parallel SRM models ($h=4, n_{ctx}=1024, b=64$, mixed heads, decay) on 2x H100.}
    \label{tables5}
    \end{table}
    \end{center}

    \begin{center}
    \begin{table}[H]
    \small
    \begin{center}
    \renewcommand{\arraystretch}{1.2}
    \begin{tabular}{|l c c c|} 
    \hline
      $d_m$ & Throughput & Memory & Parameter Count (M) \\
     \hline \hline
      128 & 8.05 & 14329 & 5.85 \\
      256 & 6.11 & 21713 & 19.03 \\
      512 & 3.78 & 34491 & 69.42 \\
      1024 & 1.82 & 63295 & 252.29 \\
      2048 & *0.693 & *170018 & 974.30 \\
      \hline 
    \end{tabular}
    \end{center}
    \vspace{0.1cm}
    \caption{Training Throughput (samples/sec) and Memory (GB per device) for Mamba (2) models ($h=8, n_{ctx}=512, b=128$) on 2x H100. *with gradient accumulation}
    \label{tables6}
    \end{table}
    \end{center}

    \begin{center}
    \begin{table}[H]
    \small
    \begin{center}
    \renewcommand{\arraystretch}{1.2}
    \begin{tabular}{|l c c c|} 
    \hline
      $d_m$ & Throughput & Memory & Parameter Count (M) \\
     \hline \hline
      128 & 6.55 & 19193 & 5.85 \\
      256 & 5.08 & 25757 & 19.03 \\
      512 & 3.34 & 67733 & 69.42 \\
      1024 & 1.70 & 67733 & 252.29 \\
      2048 & *0.676 & *167570 & 974.30 \\
      \hline 
    \end{tabular}
    \end{center}
    \vspace{0.1cm}
    \caption{Training Throughput (samples/sec) and Memory (GB per device) for Mamba (2) models ($h=8, n_{ctx}=1024, b=64$) on 2x H100. *with gradient accumulation}
    \label{tables7}
    \end{table}
    \end{center}

\end{document}